\documentclass{article}

\usepackage[preprint,nonatbib]{neurips_2024}

\usepackage[utf8]{inputenc} 
\usepackage[T1]{fontenc}    
\usepackage{hyperref}       
\usepackage{url}            
\usepackage{booktabs}       
\usepackage{amsfonts}       
\usepackage{nicefrac}       
\usepackage{microtype}      
\usepackage{xcolor}         
\usepackage{xspace}
\usepackage{enumitem}
\usepackage{graphicx}
\usepackage{bm}
\usepackage{multirow}
\usepackage{mathtools}
\usepackage{subfigure}
\usepackage{wrapfig}

\newcommand{\eg}{\emph{e.g.}}
\newcommand{\ie}{\emph{i.e.}}

\definecolor{gray}{RGB}{120, 120, 120}
\definecolor{darkred}{RGB}{135, 12, 12}
\definecolor{lightblue}{RGB}{0, 128, 192}

\title{ViD-GPT: Introducing GPT-style Autoregressive Generation in Video Diffusion Models}

\author{
    Kaifeng Gao$^{1*}$, \;
    Jiaxin Shi$^{2}$\thanks{Equal contribution. The work was done when Kaifeng Gao served as a research intern at Huawei Cloud Computing.}, \;
    Hanwang Zhang$^{3}$, \;
    Chunping Wang$^{4}$, \;
    Jun Xiao$^{1}$ \;
    \\
    $^{1}$Zhejiang University ~~ $^{2}$Huawei Cloud Computing \\
    $^{3}$Nanyang Technological University ~~ $^{4}$Finvolution Group \\
 kite\_phone@zju.edu.cn ~shijiaxin3@huawei.com ~hanwangzhang@ntu.edu.sg \\
 wangchunping02@xinye.com  ~ junx@cs.zju.edu.cn 
}

\begin{document}

\maketitle

\begin{abstract}
  With the advance of diffusion models, today's video generation has achieved impressive quality. But generating temporal consistent long videos is still challenging. A majority of video diffusion models (VDMs) generate long videos in an autoregressive manner, \ie, generating subsequent clips conditioned on last frames of previous clip. However, existing approaches all involve \emph{bidirectional} computations, which restricts the receptive context of each autoregression step, and results in the model lacking long-term dependencies. Inspired from the huge success of large language models (LLMs) and following GPT (generative pre-trained transformer), we bring causal (\ie, unidirectional) generation into VDMs, and use past frames as prompt to generate future frames. For \textbf{Causal Generation}, we introduce causal temporal attention into VDM, which forces each generated frame to depend on its previous frames. For \textbf{Frame as Prompt}, we inject the conditional frames by concatenating them with noisy frames (frames to be generated) along the temporal axis. Consequently, we present \textbf{Vi}deo \textbf{D}iffusion \textbf{GPT} (ViD-GPT). Based on the two key designs, in each autoregression step, it is able to acquire long-term context from prompting frames concatenated by all previously generated frames. Additionally, we bring the kv-cache mechanism to VDMs, which eliminates the redundant computation from overlapped frames, significantly boosting the inference speed. Extensive experiments demonstrate that our ViD-GPT achieves state-of-the-art performance both quantitatively and qualitatively on long video generation. Code is available at \url{https://github.com/Dawn-LX/Causal-VideoGen}.

\end{abstract}

\section{Introduction}\label{sec:introduction}

\begin{figure}
  \centering
  \includegraphics[width=\linewidth]{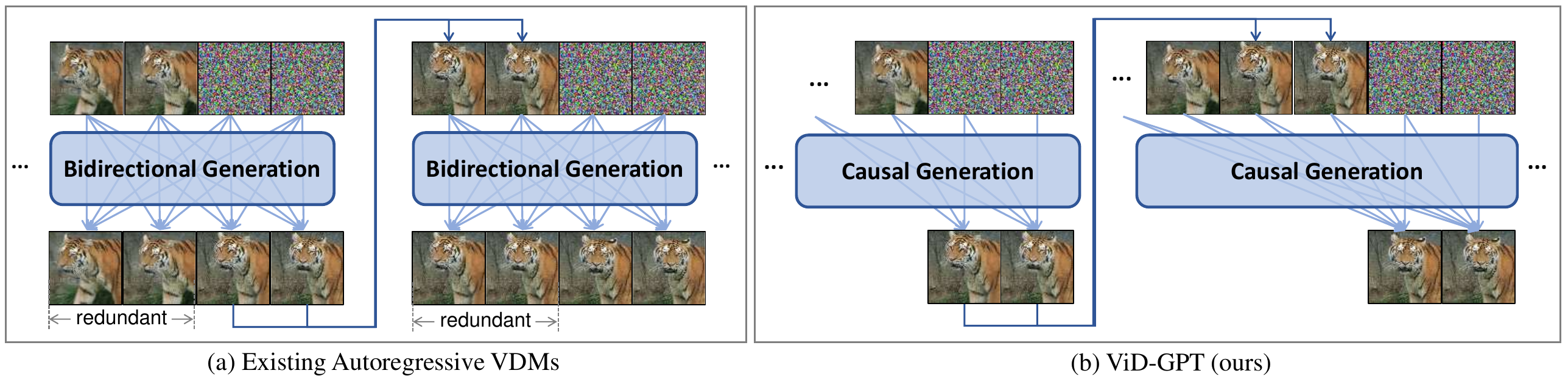}
  \caption{(a):Existing autoregressive VDMs. They generate a subsequent chunk (we draw a 4-frame chunk for brevity) conditioned on last frames of the previous chunk, with an overlapping length of half a chunk (\eg, \cite{henschel2024streamingt2v,wang2023gen}).
  (b) Our ViD-GPT. We use causal generation conditioned on all previous frames, and eliminate the redundant computation of overlapped frames by kv-cache (\emph{cf}. Figure~\ref{fig:pipeline}).}\label{fig:fig1} 
  \vspace{-2ex}
\end{figure}

The vision community has witnessed significant achievements in text-to-video (T2V) generation~\cite{khachatryan2023text2video,blattmann2023align,guo2024animatediff,chen2023seine,girdhar2023emu,ma2024latte}  due to the rise of diffusion-based image generation models~\cite{rombach2022high,peebles2023scalable,chen2024pixart}. Although excellent video quality is achieved, the majority focuses on short video generation (typically 16 frames), limiting their applications in real-world scenarios such as movie production. To this end, recent works~\cite{he2022latent,wang2023gen,dai2023animateanything,zeng2023make,lu2024vdt,henschel2024streamingt2v} have turned their attention to long-video generation. They typically train a video diffusion model (VDM) conditioned on past video frames, to generate future frames. Then, long videos can be extrapolated by autoregressively generating subsequent clips conditioned on last frames of previous clip, as illustrated in Figure~\ref{fig:fig1}~(a).

However, existing VDMs all involve \emph{bidirectional} computations during video generation (\eg, temporal attention or temporal convolution layers). 
Due to the bidirectional computation, the length of each autoregression chunk must be the same as the training video. For example, a VDM trained on 16-frame videos can only generate a 16-frame chunk at each autoregression step, where each chunk consists of, \eg, 8 conditional frames from previous chunk (as the context), and 8 frames to be denoised. Extending the chunk length to acquire context of further previous chunks violates the alignment to training and leads to incorrect feature computation. This inherently results in the model lacking long-term dependencies. Because of the limited context, the generation results often exhibit undesired object appearance changes or quality degeneration (\emph{cf}. Figure~\ref{fig:qualitative}), and periodic content mutations at the junctions of autoregression chunks (\emph{cf}. Figure~\ref{fig:FD}). 

In fact, bidirectional computation contradicts the inherent characteristics of long video generation, \ie, the model should not be aware of what comes next during the autoregression process. If we turn the attention to natural language generation, existing large language models (LLMs) all use unidirectional computation (\ie, causal and decoder-only Transformer structures as GPTs~\cite{radford2018improving,radford2019language,brown2020language}) to generation extremely long token sequences. 
Based on the above considerations, we propose \textbf{Vi}deo \textbf{D}iffusion \textbf{GPT} (ViD-GPT), a long video generation paradigm
which introduces GPT-style autoregressive generation into video diffusion models (VDMs), as shown in Figure~\ref{fig:fig1}~(b). 

To bring the GPT-style autoregression into video diffusion models, we introduce two key designs: 1) \textbf{Causal Generation}. We modify the temporal attention in existing VDMs to causal attention and build a causal VDM. It ensures each generated frame is only depended on its previous frames. 2) \textbf{Frame as Prompt}. We treat the conditional frames as prompt and concatenate them with noisy frames (frames to be generated) along temporal axis. That is, the conditional frames are treated as prompt tokens during all the sequence operations in the causal VDM.
Based on the above designs, we train and inference our model as follows. At the training stage, we randomly keep $P$ frames unnoised (as prompt), and 
let model denoise the remaining frames (Sec.~\ref{sec:training}). Due to the causal generation, the model is forced to denoise each frame without ``seeing'' its future frames. 
Then, at the inference stage, the prompt frames can be gradually extended as the autoregression proceeds. It starts from a given first frame, and autoregresssively generates $n$ frames in each step, conditioned on prompt frames concatenated by all previous frames.
Furthermore, we introduce the kv-cache mechanism prevalent in today's LLMs into the causal VDM (Sec.~\ref{sec:inference}). It eliminates the redundant computation from overlapped frames and significantly boosts the inference speed.


We evaluate our ViD-GPT quantitatively on two public datasets MSR-VTT~\cite{xu2016msr} and UCF-101~\cite{soomro2012ucf101}. It achieves state-of-the-art Fr\'echet Video Distance (FVD) scores in both zero-shot and finetuning settings (\emph{cf}. Table~\ref{tab:zeroshot}~\&~\ref{tab:finetune}). We also compare ViD-GPT with existing autoregressive VDMs on qualitative examples (\emph{cf}. Figure~\ref{fig:qualitative}). The results show that our model is more robust for content mutations and quality degeneration during the long video generation. 


In summary, we make three contributions in this paper: 1) We present a GPT-style autoregression paradigm for long video generation, introducing causal generation and frame as prompt mechanisms into video diffusion models (VDMs). 
2) To the best of our knowledge, we are the first one to introduce kv-cache mechanism into video generation on VDMs. It is verified to significantly boost the inference speed (\emph{cf}. Table~\ref{table:fps}).
3) Our ViD-GPT achieves state-of-the-art performance both quantitatively on public benchmarks and qualitatively on long video generation results.

\section{Related Work}
\subsection{Video Diffusion Models}

With the success of diffusion based methods~\cite{ho2020denoising,nichol2021improved,rombach2022high,peebles2023scalable} in image generation applications, a majority of studies turn their attention to video diffusion models (VDMs). 
Some works\cite{lu2023flowzero,khachatryan2023text2video,hong2023large,zhang2023controlvideo} develop training-free methods for zero-shot video generation based on pretrained image diffusion models (\eg, Stable Diffusion~\cite{rombach2022high}). To leverage video training data and improve the generation quality, many works\cite{ge2023preserve,guo2024animatediff,wang2023modelscope,ren2024consisti2v,dai2023animateanything} extend the 2D Unet in text-to-image diffusion models with temporal attention layers or temporal convolution layers. Recent studies~\cite{ma2024latte,lu2024vdt} also build VDMs based on spatial-temporal Transformers due to their inherent capability of capturing long-term temporal dependencies. Following prior works~\cite{ma2024latte,lu2024vdt}, we build our ViD-GPT based on spatial-temporal Transformers and use causal temporal attention in each Transformer block.

\subsection{Long Video Generation}

Intuitively, long video generation can be achieved by extending existing short VDMs in a training-free manner, \eg, by initializing noise sequence based on the DDIM inversion~\cite{song2021denoising,mokady2023null} of previously generated frames~\cite{oh2023mtvg}, co-denoising overlapped short clips~\cite{wang2023gen}, or iteratively denoising short clips with noise-rescheduling~\cite{qiu2024freenoise}. However, their generation quality is upperbounded by the pretrained short VDM, and the lack of finetuning also results in unsatisfied temporal consistency between short clip transitions. 

To enhance generation quality and temporal consistency, many recent studies concentrate on training autoregressive VDMs. They generate subsequent clips conditioned on last frames of previous clip\footnote{Image-to-video models can be seen as a special case of this type, which is only conditioned on one frame.}, as illustrated in Figure~\ref{fig:fig1} and discussed in Sec.~\ref{sec:introduction}. Recent works of autoregressive VDMs have studied a variety of design choices for injecting the conditioned frames, such as via adaptive layer normalization~\cite{Voleti2022MCVD,lu2024vdt}, cross-attention~\cite{zhang2023i2vgen,lu2024vdt,henschel2024streamingt2v}, and temporal-~\cite{harvey2022flexible,lu2024vdt} or channel-wise~\cite{chen2023seine,girdhar2023emu,zeng2023make} concatenation to the noisy latents. Some works~\cite{weng2023art,guo2023sparsectrl} also inject past frames by adapter-like subnets (e.g., T2I-adapter~\cite{mou2024t2i} or ControlNet~\cite{zhang2023adding}). In contrast to existing autoregressive VDMs, our ViD-GPT conducts causal generation during the autoregression process. There are also several works~\cite{ge2022long,villegas2022phenaki} which introduce causal attention for video generation. But they are all GAN-based~\cite{goodfellow2014generative}, which have been proved having worse performance than diffusion models on image and video synthesis~\cite{dhariwal2021diffusion,he2022latent,lu2024vdt}. Instead, we are the first one to introduce causal autoregressive video generation into diffusion models.

\section{Method}

\begin{figure}
  \centering
  \includegraphics[width=\linewidth]{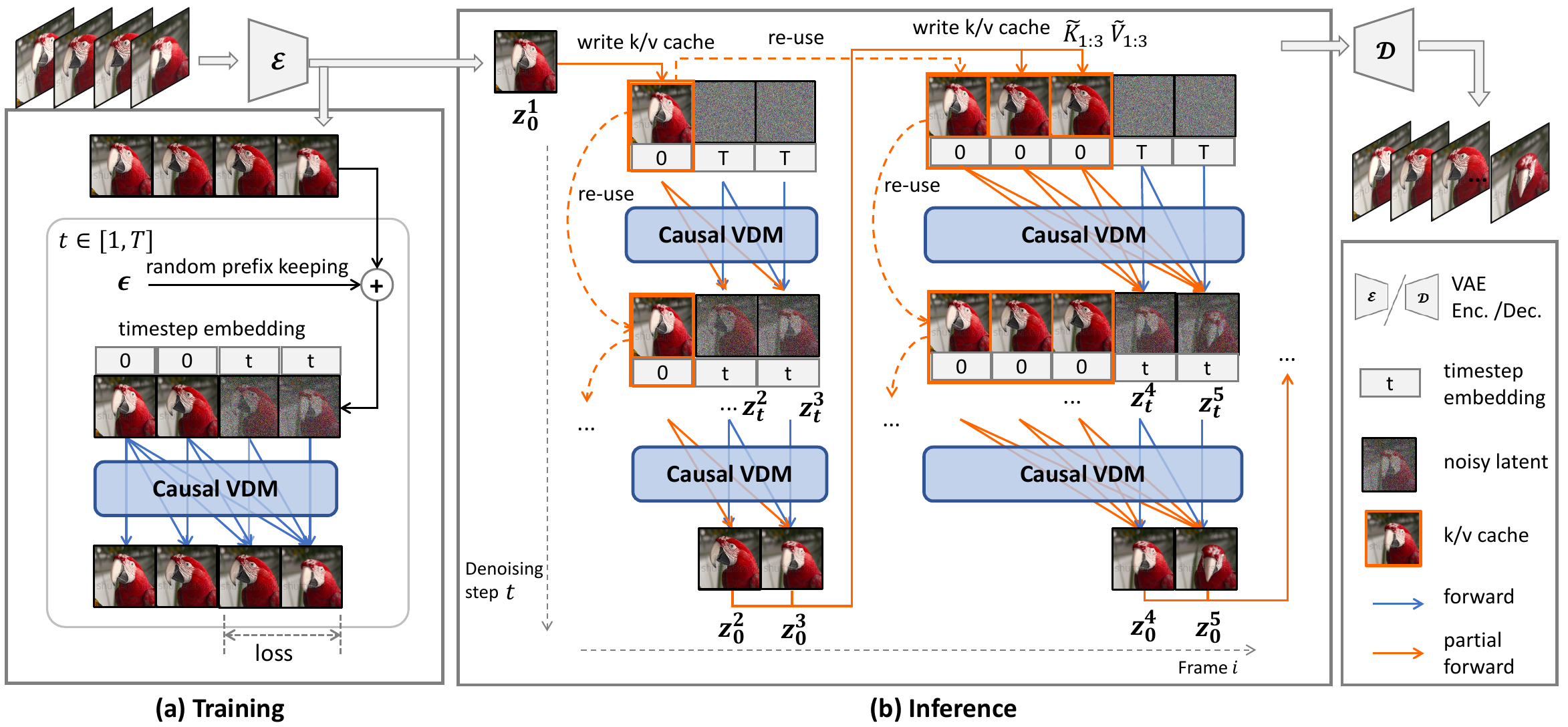}
  \caption{The overall pipeline of ViD-GPT. At the training stage (a), we randomly keep prefix frames unnoised as prompt and let the model denoise remaining frames. At the inference stage (b), the prompt frames are gradually extended (concatenated by a given first frame and all previous predictions) as the autoregression proceeds. We also introduce kv-cache (orange arrow). The model read prompt frames from kv cache without recomputing, and write new kv-cache based on the denoised latents.}\label{fig:pipeline} %
\end{figure}

We first briefly introduce the preliminaries for diffusion models and formally define the problem of autoregressive video generation (Sec.~\ref{sec:preliminaries}). Then, we introduce the VID-GPT framework, as illustrated in Figure~\ref{fig:pipeline}, in which we first build a causal VDM (Sec.~\ref{sec:causalVDM}), and then design the training stage with frame as prompt (Sec.~\ref{sec:training}) and the inference stage with kv-cache (Sec.~\ref{sec:inference}).

\subsection{Preliminaries and Problem Formulation}\label{sec:preliminaries}

\textbf{Diffusion Models}. Diffusion Models~\cite{sohl2015deep,ho2020denoising} are generative models which model a target distribution $\bm{x}_0 \sim q(\bm{x}) $ by learning a denoising process with arbitrary noise levels. To do this, firstly a diffusion process is defined to gradually corrupt $\bm{x}_0$ with Gaussian noise. Each diffusion step is $q(\bm{x}_t | \bm{x}_{t-1}) = \mathcal{N}(\bm{x}_t;\sqrt{1-\beta_t}\bm{x}_{t-1},\beta_t\bm{I})$, where $t=1,\ldots,T$ and $\beta_t \in (0,1)$ is the variance schedule. By applying the reparameterization trick~\cite{ho2020denoising}, each $\bm{x}_t$ can be sampled as $\bm{x}_t = \sqrt{\bar{\alpha}_t}\bm{x}_0 + \sqrt{1-\bar{\alpha}_t}\bm{\epsilon}_t$, where $\bm{\epsilon}_t \sim \mathcal{N}(\bm{0},\bm{I})$ and $\bar{\alpha}_t = \prod_{i=1}^t(1-\beta_i)$. 
Given the diffusion process, a diffusion model is then trained to approximate the reverse process (denoising process). Each denoising step is parameterized as $p_\theta(\bm{x}_{t-1} | \bm{x}_t) = \mathcal{N}(\bm{x}_{t-1};\bm{\mu}_\theta(\bm{x}_t,t),\bm{\Sigma}_\theta(\bm{x}_t,t))$, where $\theta$ contains learnable parameters.

\textbf{Latent Diffusion Models}. 
To reduce computational demands, latent diffusion models (LDMs)~\cite{rombach2022high} propose to modeling the diffusion-denoising process in latent space instead of the raw pixel space. This is achieved by using a pretrained variational autoencoder (VAE) $\mathcal{E}$ to compress $\bm{x}_0$ into a lower-dimensional latent representation $\bm{z}_0 = \mathcal{E}(\bm{x}_0)$. Consequently, the diffusion and denoising processes become $q(\bm{z}_t | \bm{z}_{t-1})$ and $p_\theta(\bm{z}_{t-1} | \bm{z}_t)$, respectively. The denoised latent $\hat{\bm{z}}_0$ is decoded back to the pixel space by a pretrained VAE decoder, \ie, $\hat{\bm{x}}_0 = \mathcal{D}(\hat{\bm{z}}_0)$.

Prevailing LDMs are trained with the variational lower bound of the log-likelihood of $\bm{z}_0$, reducing to $\mathcal{L}_{\text{vlb}}(\theta) = -p_\theta(\bm{z}_0 | \bm{z}_1) + \sum_{t} D_{KL}(q(\bm{z}_{t-1} | \bm{z}_t, \bm{z}_0) \| p_\theta(\bm{z}_{t-1} | \bm{z}_t))$. Since $q$ and $p_\theta$ are Gaussian, the $D_{KL}$ term is determined by the mean $\bm{\mu}_\theta$ and the covariance $\bm{\Sigma}_\theta$. By reparameterizing $\bm{\mu}_\theta$ as a noise prediction network $\bm{\epsilon}_\theta$ and fixing $\bm{\Sigma}_\theta$ as a constant variance schedule~\cite{ho2020denoising}, the model can be trained using a simplified objective $\mathcal{L}_{\text{simple}}(\theta)= \mathbb{E}_{\bm{z},\bm{\epsilon},t}\left[ \|\bm{\epsilon}_\theta(\bm{z}_t,t) - \bm{\epsilon}_t\|_2^2\right]$. Following Nichol and Dhariwal~\cite{nichol2021improved}, we can also train the LDM with learnable covariance $\bm{\Sigma}_\theta$, by optimizing the full $D_{KL}$ term (\ie, training with the full $\mathcal{L}_{\rm vlb}$), where $\bm{\epsilon}_\theta$ and $\bm{\Sigma}_\theta$ implemented using the same denoising network (denoted as $\bm{\epsilon}_\theta(\bm{z}_t,t)$ for brevity) which predicts both the noise and covariance~\cite{peebles2023scalable,ma2024latte}. 

\textbf{Problem formulation}. 
In this paper, we aim to achieve autoregressive video generation based on latent video diffusion models (VDMs). Let $\bm{z}_0^{1:N} = [\bm{z}_0^1,\dots, \bm{z}_0^N] \in \mathbb{R}^{N \times H \times W \times C}$ be the latent sequence encoded by a pretrained VAE, where $N$ is the number of frames, $H\times W$ is the down sampled resolution, and $C$ is the number of channels. Our goal is to learn a video diffusion model in the latent space: $p_\theta(\bm{z}_0^{1:N} | \bm{c})$, where $\bm{c}$ denotes text conditions, and $\theta$ is implemented by a denoising network, modeled as $\bm{\epsilon}_\theta(\bm{z}_t^{1:N},\bm{c},t)$. In the inference stage, we generate the video autoregressively with a step length of $n$ frames. Each autoregression step consists $T$ denoising steps, in which each denoising step samples $\bm{z}_{t-1}^{k:k+n} \sim p_\theta(\bm{z}_{t-1}^{k:k+n}|\bm{z}_t^{k:k+n},\bm{z}_0^{1:k},\bm{c})$, \ie, conditioned on the text prompt and previously generated frames. Note that the first autoregression step starts from a given image.

\subsection{Causal Video Diffusion Models}\label{sec:causalVDM}

\textbf{Model Architecture}. We build our ViD-GPT using spatial-temporal Transformer~\cite{peebles2023scalable,ma2024latte,lu2024vdt} as the foundation model due to its superior sequence modeling capabilities. 
It contains a embedding layer followed by a stack of spatial-temporal Transformer blocks, and a final feed-forward network (FFN). In the embedding layer,  the latent code $\bm{z}_t^{1:F}$ is first divided into non-overlapping patches 
and then embedded by linear projection. Following Vision Transformer (ViT)~\cite{dosovitskiy2020image}, each patch is added with sinusoidal spatial and temporal positional embeddings. In the spatial-temporal Transformer blocks, each block consists of a spatial-attention, a temporal-attention, a cross-attention, and an FFN, with residual connections and adaptive layer normalization (AdaLN). Following~\cite{ma2024latte,chen2024pixart}, we inject the diffusion step ($t$) embeddings via AdaLN and add text guidance via cross-attention. The final FFN decode the patch embeddings to the noise and covariance prediction at each diffusion step $t$. The detailed model structure is left in the Appendix (Sec.~\ref{app:hyperparameters})

\textbf{Causal Attention}. The key design for introducing causality to the VDM lies in the causal temporal attention: We mask the attention map to force each frame only attend to its previous frames. Specifically, let $\Tilde{\bm{z}}_t^{1:N} \in \mathbb{R}^{N \times H \times W \times C'}$ be the latent code\footnote{The spatial resolution is actually down sampled for the patchified embeddings. Here we slightly abuse the notation and still use $H\times W$ without introducing ambiguity. We do not use temporal patchy.} input to each causal attention layer. 
$\Tilde{\bm{z}}_t^{1:N}$ is first permuted by treating $H\times W$ as the batch dimension, and then linearly projected to query, key, and value features as $\bm{Q,K,V}\in \mathbb{R}^{N \times C''}$ (for each spatial grid in a batch). The causal attention is computed as (we only describe one attention head and omit the diffusion step $t$ for brevity):
\begin{align}\label{Eq_selfattn}
  \text{CausalAttn}(\bm{Q,K,V}) = \text{Softmax}\left(\bm{QK}^{\rm T} / \sqrt{C''} + \bm{M}\right)\bm{V}, %
\end{align}
where $\bm{M} \in \mathbb{R}^{N\times N}$ is the attention mask with $\bm{M}_{i,j}=-\infty$ if $i < j $ else $0$. 
It is worth noting that although we choose spatial-temporal Transformer as our foundation model, our ViD-GPT paradigm is not bonded with the Transformer structure. It can also support UNet-based VDMs~\cite{he2022latent,guo2024animatediff,ren2024consisti2v,zeng2023make} with temporal attention layers, \ie, by replacing the full attention as our causal attention.

\subsection{Training with Frame as Prompt}\label{sec:training}


In the training stage, the model learns to denoise the noisy latent sequence. In contrast to conventional VDMs with bidirectional temporal attention, the causal VDM propagates the information unidirectionally. Thus, if we simplly applying causal attention, each noisy latent will only receive information from its predecessor latents (which is also noisy), and it might be suboptimal for long video generation (\emph{cf}. Sec.~\ref{sec:abaltion}). This encourages us to add clean latent frames as prompt to guide the denoising process. Particularly, we randomly keep prefix frames unnoised as prompt, and use the timestep embeddings of $t=0$ for them, as shown in Figure~\ref{fig:pipeline}~(a). Since the model only partially deonises the latent sequence, we compute the loss excluding the prompt portion. To be specific, let $P$ be length of prompt. At each training iteration with diffusion step $t$, the loss is computed as 
\begin{align}
    \widetilde{\mathcal{L}}_{\text{simple}}(\theta)= \mathbb{E}_{\bm{z},\bm{\epsilon},t}\left[ \|
    (\bm{\epsilon}_\theta([\bm{z}_0^{1:P},\bm{z}_t^{P+1:N}],\bm{c},t) - \bm{\epsilon}_t) \odot \bm{m}
    \|_2^2\right],
\end{align} 
where $[\cdot,\cdot]$ stands for concatenating along temporal axis, and $\bm{m} \in \{0,1\}^{N}$ is a binary loss mask with $\bm{m}_i=1$ if $i > P$ else $0$. We learn the covariance $\bm{\Sigma}_\theta$ by the full $\widetilde{\mathcal{L}}_{\text{vlb}}$ (applied with the same loss mask $\bm$). Following Nichol and Dhariwal~\cite{nichol2021improved} and prior works~\cite{peebles2023scalable,ma2024latte}, the model is optimized by a combined loss of $\widetilde{\mathcal{L}}_{\text{simple}} + \widetilde{\mathcal{L}}_{\text{vlb}}$.

\textbf{Frame Prompt Enhancement}. The guidance of frame prompt usually gradually degrades during the propagating through causal temporal attention. Inspired by prior works~\cite{hu2023animate,ren2024consisti2v}, we inject extra reference through spatial attention layers to enhance the guidance. In detail, let $\bar{\bm{z}}_t^{1:N} \in \mathbb{R}^{N\times H\times W \times C'}$ be the input of each spatial attention layer. Here the number of frames $N$ is treated as batch dimension and $ H\times W $ is flattened for the attention operation. For $\bar{\bm{z}}_t^{i}$ of $i$-th frame, the query, key and value are:
\begin{align}
    \bm{Q}_s = \bm{W}^Q\bar{\bm{z}}_t^{i}, \bm{K}_s = \bm{W}^K\bar{\bm{z}}_t^{i}, \bm{V}_s = \bm{W}^V\bar{\bm{z}}_t^{i}, ~\bm{Q}_s,\bm{K}_s,\bm{V}_s \in \mathbb{R}^{(H\times W)\times C'},
\end{align}
where $\bm{Q}_s,\bm{K}_s,\bm{V}_s$ are learnable projection matrices. 
To enhance the guidance of frame prompt, we concatenate a sub-prompt of length $P'$ to the noisy latent on the spatial dimension, and repeat the clean latent with itself by $P'$ times (\emph{cf}. Figure~\ref{fig:kvcache2}~(a)). The enhanced key is computed as 
\begin{align}
    \bm{K}'_s = \bm{W}^K[\bar{\bm{z}}_t^i;\bar{\bm{z}}_0^{P-1};...;\bar{\bm{z}}_0^{P-P'}], ~(\text{for} ~i > P); 
    \bm{K}'_s = \bm{W}^K[\bar{\bm{z}}_0^i;...;\bar{\bm{z}}_0^i] ~(\text{for} ~i \leq P), 
\end{align}
where $[\cdot ;\cdot]$ stands for concatenating on spatial dimension and $\bm{K}_s \in \mathbb{R}^{((P'+1)\times H\times W)\times C'}$. We do the same operation to obtain the enhanced value $\bm{V}'_s$. Finally, the enhanced spatial attention is
\begin{equation}
    \text{Attention} (\bm{Q}_s, \bm{K}'_s,\bm{V}'_s) = \text{Softmax} ( \bm{Q}_s \bm{K}_s^{\rm \prime T} / \sqrt{C''})\bm{V}'_s
\end{equation}
We empirically show the frame prompt enhancement helps the model alleviate the quality degeneration during autoregressive long video generation (\emph{cf}. Sec.~\ref{sec:abaltion}). 

\subsection{Inference Boosted with KV-cache}\label{sec:inference}

\begin{figure}
    \vspace{-3ex}
    \centering
    \includegraphics[width=\linewidth]{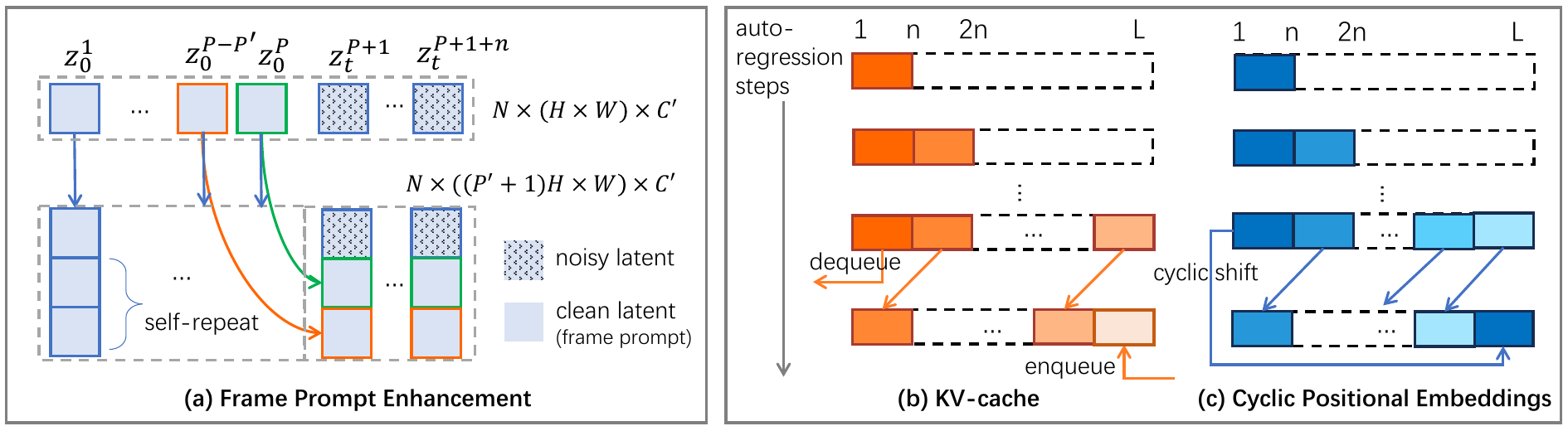}
    \caption{Illustration of frame prompt enhancement (a), KV-cache (b), and cyclic positional embeddings (c). Here $n$ is the length of each autoregression chunk. $L$ is a predefined length. The model starts dequeuing kv-cache and shifting positional embedding when the length of video exceeds $L$. }\label{fig:kvcache2} %
    \vspace{-2ex}
\end{figure}

In the inference stage, the model generate videos autoregressively. At each autoregression step $i$, it aims to predict $n$ frames given the previously generated sequence $\bm{z}_0^{1:k}$ as frame prompt, where $k$ is the number of accumulated generated frames. Each autoregression step goes through a $T$-step denoising process, \ie, the VDM is called $T$ times with a latent sequence of length $k+n$. To avoid the redundant computation of previously generated frames, we introduce a kv-cache mechanism inspired from large language models (LLMs), as shown in Figure~\ref{fig:pipeline}~(b). Specifically, let $\Tilde{\bm{K}}_{1:k},\Tilde{\bm{V}}_{1:k} \in \mathbb{R}^{k\times C''}$ (considering only one spatial grid) be the accumulated cached key and value features. In the current autoregression step, the model takes as input the latent chunk $\bm{z}_t^{k:k+n}$ and computes the query, key, and value features as $\bm{Q}_{k:k+n}, \bm{K}_{k:k+n},\bm{V}_{k:k+n} \in \mathbb{R}^{n\times C''}$. It first concatenates the key and vaule with the cached ones, \ie,
\begin{align}\label{eq:kvcache}
    \bm{K}_{1:n} = [\Tilde{\bm{K}}_{1:k},\bm{K}_{k:k+n}],~
    \bm{V}_{1:n} = [\Tilde{\bm{V}}_{1:k},\bm{V}_{k:k+n}],~
    \bm{K}_{1:n}, \bm{V}_{1:n} \in \mathbb{R}^{(k+n) \times C''},
\end{align}
and then computes the causal temporal attention  as 
\begin{align}\label{eq:kvcacheAttn}
    \text{CausalAttn}(\bm{Q}_{k:k+n},\bm{K}_{1:n},\bm{V}_{1:n}) = \text{Softmax}\left(\bm{Q}_{k:k+n}\bm{K}_{1:n}^{\rm T} / \sqrt{C''} + \bm{M}'\right)\bm{V}_{1:n},
\end{align}
where the attention mask $\bm{M}'$ has the shape of $n\times k$, with each item $\bm{M}'_{i,j}=-\infty$ if $i+k < j$ else $0$ (\ie, the upper triangular part of the rightmost square sub-matrix is masked out). We demonstrate that our kv-cache mechanism significantly boosts of inference speed (\emph{cf}. Table~\ref{table:fps}).

\textbf{KV-cache Writing}. Since the model has to go through $T$ denoising steps in each autoregression step, storing kv-cache for each diffusion step costs huge memory (\ie, in the shape of $T\times k \times C''$). However, thanks to our frame as prompt design, the prompt latent is always unnoised in both training and inference. This allows us to write the kv-cache only for the ``clean'' latents after the last denoising step. That is, the $\Tilde{\bm{K}}_{1:k}$ and $\Tilde{\bm{V}}_{1:k}$ in Eq.~(\ref{eq:kvcache}) are computed from previously denoised latents $\bm{z}_0^{1:k}$, and the current autoregression step will write kv cache based on the denoised latent chunk $\bm{z}_0^{k:k+n}$.

\textbf{KV-cache Dequeuing and Cyclic Positional Embeddings}. 
To enable longer video generation and save the memory storage, we dequeue the oldest kv-cache when the kv-cache to be written exceeds a predefined length $L$ (\emph{cf}. Figure~\ref{fig:kvcache2}~(b)). When the video is very long, this is fine in most cases because the oldest frames contribute little information to the latest generation chunk. We also introduce cyclic temporal positional embeddings, as shown in Figure~\ref{fig:kvcache2}~(c). It conducts a cyclic shift operation on the embeddings when the indexed position exceeds $L$. This allows the model to extend the positional embeddings with arbitrary length beyond the training length.

\section{Experiments}

\subsection{Implementation Details}\label{sec:ImplementationDetails}
We buid ViD-GPT using the spatial-temporal Transformer structure following~\cite{ma2024latte,chen2024pixart}, and initialize it from the released weight of OpenSORA~\cite{opensora}. Following PixArt-$\alpha$~\cite{chen2024pixart}, we use T5 (Flan-T5-XXL)~\cite{raffel2020exploring} as the text encoder and use the pretrained VAE from StableDiffusion~\cite{rombach2022high} as the image encoder. We train our model on a large scale video-text dataset InternVid~\cite{wang2023internvid}, by filtering it to a sub-set of 4.9M high-quality video-text pairs. We leave the training details and hyperparameters in Sec.~\ref{app:hyperparameters}.

\subsection{Comparisons for Short Video Generation}\label{sec:compareshort}

\textbf{Datasets and Evaluation Metrics}. We use two public datasets MSR-VTT~\cite{xu2016msr} and UCF-101~\cite{soomro2012ucf101}, and report Fr\'echet Video Distance (FVD)~\cite{unterthiner2019fvd} following previous works~\cite{zeng2023make,ge2023preserve,chen2023seine}. For MSR-VTT, we use the official test split which contains 2990 videos, with 20 manually annotated captions for each video. Following prior works~\cite{ren2024consisti2v,zeng2023make} and for fair comparison, we randomly select a caption for each video and generate 2990 videos with resolution $16 \times 256 \times 256$ for evaluation. For UCF-101, as it only contains label names, we employ the descriptive text prompts from PYoCo~\cite{ge2023preserve}, and generate 2048 samples with uniform distribution for each category following previous works~\cite{he2022latent,ge2023preserve}. More implementation details of these metrics are left in Appendix Sec.~\ref{app:eval}.


\begin{figure}
    \vspace{-3ex}
    \centering
    \includegraphics[width=\linewidth]{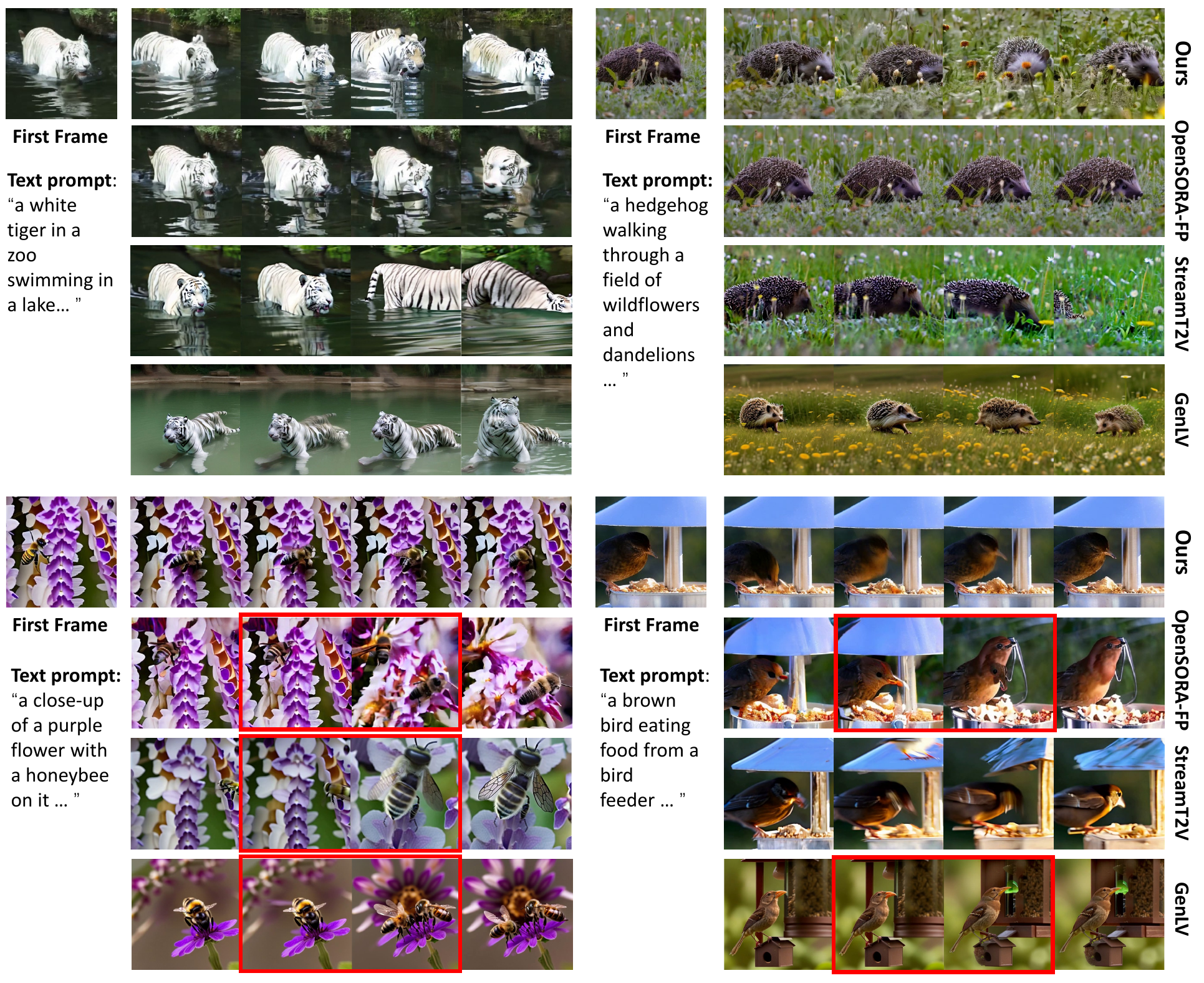}
    \caption{Qualitative comparisons between GenLV~\cite{wang2023gen}, StreamT2V~\cite{henschel2024streamingt2v}, OpenSORA~\cite{opensora} combined with our frame as prompt (OpenSORA-FP), and our ViD-GPT. For the top two examples, we sample 32 frames with an interval of 8 frames. For the bottom two examples, we select representative consecutive clips, where mutations between consecutive frames are highlighted with red boxes. Refer to Figure~\ref{fig:qualitativeApp} for more details. Note that GenLV is conditioned only on the text prompt.
    }\label{fig:qualitative} %
    \vspace{-3ex}
\end{figure}

\textbf{Quantitative Results}. We evaluate the short video generation quality of ViD-GPT and compare it to state-of-the-art text-to-video (T2V) generation methods: ModelScope~\cite{wang2023modelscope}, VideoComposer~\cite{wang2023videocomposer}, Video-LDM~\cite{blattmann2023align}, PYoCO~\cite{ge2023preserve}, and Make-A-Video~\cite{singer2022make}. To align with our setting, we also compare ViD-GPT against image conditioned T2V models: AnimateAnything~\cite{dai2023animateanything}, PixelDance~\cite{zeng2023make}, and SEINE~\cite{chen2023seine}. We evaluate the zero-shot performance on MSR-VTT and UCF-101, and report the FVD scores in Table~\ref{tab:zeroshot}. We also finetune our model on UCF-101 training set and compare it with SOTA video generation methods MCVD~\cite{Voleti2022MCVD}, VDT~\cite{lu2024vdt}, DIGAN~\cite{yu2022generating}, TATS~\cite{ge2022long}, LVDM~\cite{he2022latent}, PVDM~\cite{yu2023video}, and Latte~\cite{ma2024latte}. We report the finetuned FVD scores in Table~\ref{tab:finetune}. The FVD results show that our ViD-GPT has comparable short video generation with SOTA methods.

\begin{table}
       \parbox{.5\linewidth}{
           \centering
           \setlength{\tabcolsep}{0.2pt}
           \caption{Zero-shot FVD performance on MSR-VTT and UCF-101 test set. All methods generate video with resolution of $16\times 256\times 256$.}
            \begin{tabular}{lccc}
            \hline
            Method          & Condition   & MSRVTT~ & UCF101 \\ \hline
            ModelScope~\cite{wang2023modelscope}      & text        & 550     & 410     \\
            VideoComposer~\cite{wang2023videocomposer}   & text         & 580     & -       \\
            Video-LDM~\cite{blattmann2023align}       & text        & -       & 550.6   \\
            PYoCo~\cite{ge2023preserve}           & text        & -       & 355.2  \\
            Make-A-Video~\cite{singer2022make}    & text        & -       & 367.2  \\ \hline
            AnimateAnything~\cite{dai2023animateanything} & text\&image & 443     & -       \\
            PixelDance~\cite{zeng2023make}      & text\&image & 381     & \textbf{242.8}   \\
            SEINE~\cite{chen2023seine}           & text\&image & \textbf{181}   & -       \\
            ours            & text\&image & \textbf{181}   & 277.7   \\ \hline
            \end{tabular}
            \label{tab:zeroshot}
        }
        \hfill
       \parbox{.4\linewidth}{
           \centering
           \caption{Finetuning setting of FVD performance on UCF-101 test set. $^*$ means trained on both train and test set.}
           \begin{tabular}{ccc}
            \hline
            Method & Res. & FVD    \\ \hline
            MCVD~\cite{Voleti2022MCVD}    & $64^2$   & 1143   \\
            VDT~\cite{lu2024vdt}     & $64^2$   & 225.7  \\ \hline
            DIGAN$^*$~\cite{yu2022generating}   & $128^2$ & 577    \\
            TATS~\cite{ge2022long}    & $128^2$ & 420    \\ 
            VideoFusion~\cite{luo2023videofusion}   & $128^2$ & 220    \\ \hline
            LVDM~\cite{he2022latent}    & $256^2$ & 372    \\
            PVDM$^*$~\cite{yu2023video}    & $256^2$ & 343.6  \\
            Latte~\cite{ma2024latte}   & $256^2$ & 333.6 \\
            Ours    & $256^2$ & \textbf{184.5} \\ \hline
            \end{tabular}
           
           \label{tab:finetune}
       }
\end{table}


\subsection{Comparisons for Long Video Generation}\label{sec:compareLong}



\textbf{Baseline Methods}. We compare ViD-GPT with three baselines: Gen-L-Video (GenLV)~\cite{wang2023gen}, StreamT2V~\cite{henschel2024streamingt2v}, and OpenSORA~\cite{opensora}. Specifically, GenLV utilizes a base model AnimateDiff~\cite{guo2024animatediff} and conducts co-denoising for overlapped 16-frame clips (we implement it with an overlapping length of 8 frames). StreamT2V extends a short VDM conditioned on previously generated frames and performs autoregressive generation for long videos. We implement StreamT2V based on Stable Video Diffusion (SVD)~\cite{blattmann2023stable} which generates 16-frame clips, and also use an overlapping length of 8 frames. OpenSORA is a recently released open source VDM built with spatial-temporal Transformers. We finetune it with our frame as prompt design (termed as OpenSORA-FP) and conduct autoregressive long video generation, where each generation chunk is 32 frames with 16 frames overlapping. 

\begin{wrapfigure}{r}{40ex}
    \vspace{-3ex}
    \centering
    \includegraphics[width=\linewidth]{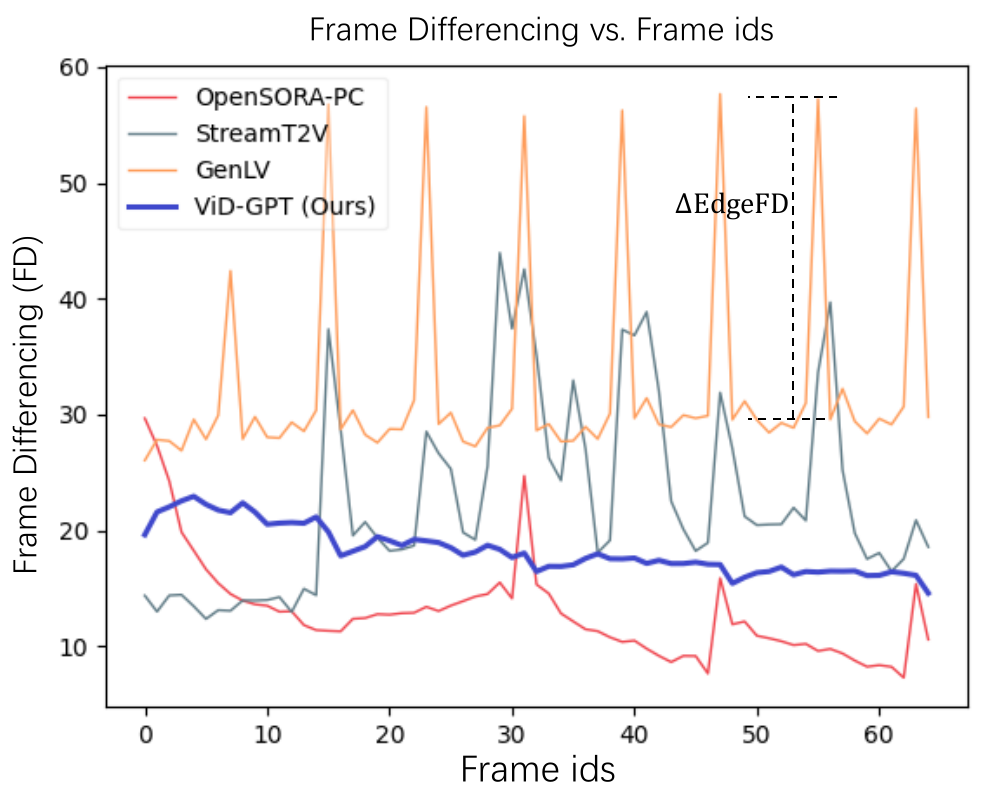}
    \vspace{-2ex}
    \caption{Frame Differecing vs. Frame ids curve on MSR-VTT test set.}\label{fig:FD}
    \vspace{-1ex}
\end{wrapfigure}

\textbf{Evaluation Metrics}. For autoregressive long video generation approaches, the video quality is determined by two key aspects: 1) \textbf{Transition consistency}, \ie, how much content mutation between chunks generated by adjacent autoregression steps. 2) \textbf{Long-term consistency}, \ie, how does object appearance/characteristic change during long-term generation. We introduce two new metrics to assess these two consistencies. For transition consistency, we introduce $\Delta$EdgeFD. We first compute the absolute pixel differences between consecutive frames, \ie, frame differencing (FD)~\cite{dai2023animateanything}. 
Then $\Delta$EdgeFD computes the incremental FD of edge frames at the junctions of autoregression chunks against the average FD in each chunk. A higher $\Delta$EdgeFD indicates more content changes in the edge frame than average. For long-term consistency, we introduce Step-FVD. It computes the FVD score of each autoregression chunk against the first chunk, which reflects long-term content changes during autoregression.

\begin{table}[t]
    \caption{
    Step-FVD and $\Delta$EdgeFD results for long video generation on MSR-VTT test set.
    }
    \label{table:stepFVD}
    \centering
    \begin{tabular}{lcccccc}
    \hline
    \multirow{2}{*}{Method} & \multicolumn{5}{c}{Step-FVD of $i$-th autoregression chunk}    & \multirow{2}{*}{$\Delta$EdgeFD} \\
                            & $i=2$     & $i=3$     & $i=4$      & $i=5$      & $i=6$      &                         \\ \hline
    GenLV+AD~\cite{wang2023gen,guo2024animatediff}            & 282.8 & 291.4 & 299.0 & 318.2 & 310.3 & 27.1                   \\
    StreamT2V+SVD~\cite{henschel2024streamingt2v,blattmann2023stable}         & 317.5 & 434.7 & 478.2 & 462.0 & 512.4 & 6.95                    \\
    OpenSORA-FP~\cite{opensora}         & 182.9 & 210.6 & \textbf{260.8} & 284.3 & 315.1 & 6.62                    \\
    Ours                    & \textbf{160.6} & \textbf{206.5} & 262.8 & \textbf{281.3} & \textbf{304.7} & \textbf{-0.29}                   \\ \hline
    \end{tabular}
\end{table}

\textbf{Quantitative Results}. We evaluate our ViD-GPT and the three baselines on MSR-VTT test set, and generate videos with resolution of $96\times256\times256$.
We report the FD-Frame\_id curve in Figure~\ref{fig:FD}, $\Delta$EdgeFD and Step-FVD in Table~\ref{table:stepFVD}. From Figure~\ref{fig:FD}, we can observe that all the three baselines have periodic content mutations (at the edge of each autoregression chunk), especially GenLV. This is because their inherent bidirectional local computation limits the model to only acquire short-term dependencies from previous one chunk, while lack of long-term dependencies. In contrast, our ViD-GPT shows a flatter FD-Frame\_id curve, which indicates more smooth transitions between autoregression chunks. The $\Delta$EdgeFD values in Table~\ref{table:stepFVD} also show that ViD-GPT has higher transition consistency. We have a near zero $\Delta$EdgeFD (\ie, -0.29), which means the edge frames have an average FD as ordinary frames at the middle part of each chunk. For Step-FVD, our ViD-GPT have relatively low values compared to the baselines, indicating better long-term consistency.

\textbf{Qualitative Results}. We present several qualitative examples in Figure~\ref{fig:qualitative}. As shown in the figure, the baseline methods show unsatisfied temporal consistency and content keeping ability. This includes 1) sudden changes of objects, as observed in the ``honeybee'' case of StreamT2V and GenLV, and the ``bird'' case of OpenSORA-FP. 2) gradually degeneration of object appearance, as evident in the ``tiger'' and ``bird'' cases of StreamT2V. 3) unnatural content at the junctions of autoregression chunks, as displayed in the ``hedgehog'' case of GenLV. In contrast, our ViD-GPT produces natural and consistent videos. We also display detailed consecutive frames in the Appendix Sec.~\ref{app:qualitive}.

\begin{wraptable}{r}{32ex}
    \vspace{-3ex}
        \caption{Inference Speed}
        \label{table:fps}
        \centering
        \addtolength{\tabcolsep}{-2.5pt}
        \begin{tabular}{lc}
            \hline
            Method            & FPS                   \\ \hline
            StreamT2V+SVD~\cite{henschel2024streamingt2v,blattmann2023stable}     & 0.53                  \\
            GenLV+AD~\cite{wang2023gen,guo2024animatediff}          & 0.73                  \\
            OpenSORA-PC~\cite{opensora}       & 0.48                  \\
            Ours w/o kv-cache & 0.22 \\
            Ours              & \textbf{0.97}         \\ \hline
        \end{tabular}
    \vspace{-4ex}
\end{wraptable}
\textbf{Comparisons for Inference Speed}. We report the infernece speed (in frame-per-second, \ie, FPS) in Table~\ref{table:fps}. The models are all tested on a single NVIDIA A800 GPU with generation resolution $256 \times 256$. It shows that our ViD-GPT has large inference speed improvement compared to the baselines (\eg, 0.97 vs. 0.53 of StreamT2V). It also verifies the significant effectiveness of our kv-cache mechanism (\ie, 0.97 vs. 0.22).


\subsection{Ablition studies}\label{sec:abaltion}

\begin{figure}
  \centering
  \includegraphics[width=\linewidth]{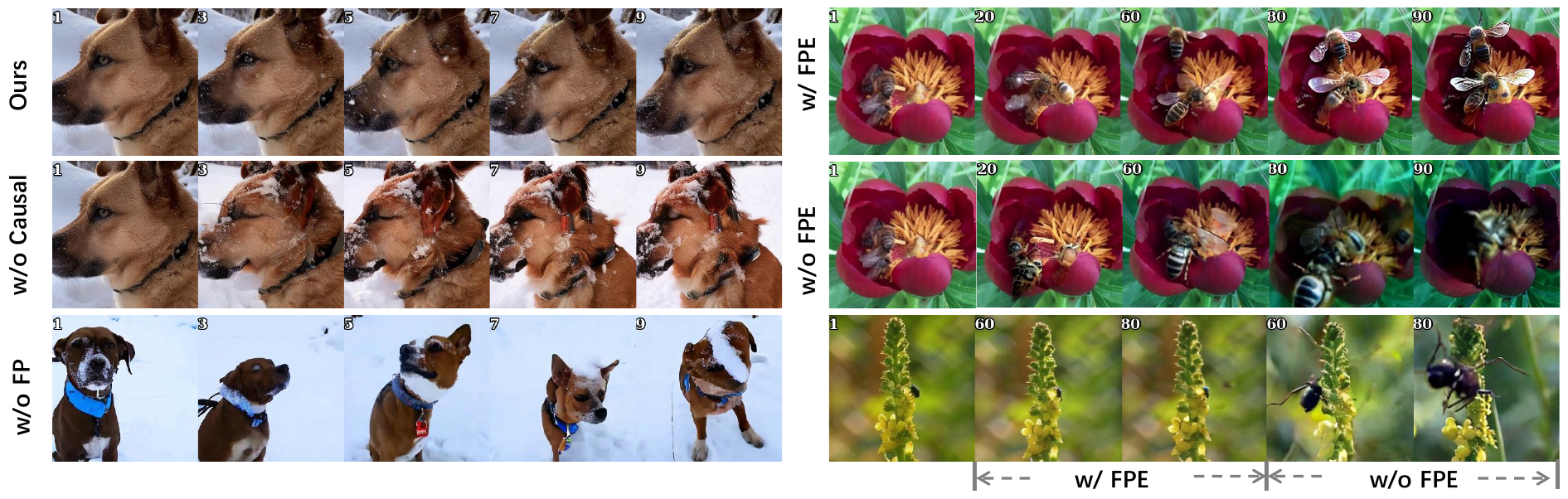}
  \caption{Qualitative ablation for causal generation and frame as prompt (FP)  (left), and for frame prompt enhancement (FPE) (right). Frame id is marked at the top-left corner in each image.}\label{fig:abaltion} %
\end{figure}

\textbf{Causal Generation and Frame as Prompt}. We conduct ablation studies to show that either causal generation or frame as prompt used along has its inherent drawbacks. The qualitative results are shown in Figure~\ref{fig:abaltion}~(left). For causal generation, we compare our model with the OpenSORA-FP baseline, which is trained with frame as prompt while without causal attention (``w/o causal'' in the middle row). We observe that the object appearance changes at the very beginning. This is because its bidirectional attention computation, which makes the noisy latents dominate the video generation and lacks the ability of following the given first frame. For removing frame as prompt (\ie, ``w/o FP'' in the bottom row), it shows that the generated frames are incoherent. This verifies that applying causal attention without frame as prompt is inappropriate, as discussed in Sec.~\ref{sec:causalVDM}. 

\textbf{Frame Prompt Enhancement}. We also conduct ablation study to verify the effectiveness of Frame Prompt enhancement (FPE), as illustrated in Figure~\ref{fig:kvcache2}~(right). The results demonstrate that FPE can improve the long-term consistency. For example, the color tone of the flower (middle row) and the appearance of the bug (bottom row) gradually change in the results of ``w/o FPE''.

\section{Conclusion and Limitations}\label{sec:conclusion}
In this paper, we present a new GPT-style autoregressive paradigm for long video generation. we introduce two key designs to video diffusion models (VDMs), \ie, causal generation and frame as prompt. Based on these designs, the mdoel is able to generate video frames autoregressively, and acquire long-term dependencies from all previous frames. We also introduce kv-cache mechanism to further boost the inference speed. Extensive experiments demonstrate the state-of-the-art long video generation performance of our ViD-GPT.

\textbf{Limitations}. 1) Due to limited computational resources, we only train our model with a relatively low resolution, \ie, $256\times 256$. 2) Our method is currently designed for image conditioned text-to-video (T2V) generation. We tried to randomly drop out the first prompt frame during training, and perform pure T2V generation autoregressively. But the results are not satisfied. 3) Our model shares some common limitations of diffusion models, \eg, unable to correctly render human faces and hands.

\bibliographystyle{plain}
{
\small
\bibliography{ReferenceBib}
}


\section{Appendix}


\subsection{Broader Impacts}\label{app:impacts}

Our ViD-GPT is a generic long video generation paradigm. It is not bounded with a certain type of video diffusion models (VDMs) and is potentially powerful to boost existing VDMs to generate high-quality long videos. The High-quality video generation techniques have a revolutionary impact on the field of content creation industry, and have great potential commercial values. Meanwhile, it's necessary to note that ViD-GPT also has the inherent risks of common image/video generation models, such as generating videos with harmful or offensive contents, or be used by malicious actors for generating fake news. We can use some watermarking technologies (\eg, \cite{lukas2023ptw}) to avoid the generated videos being abused.

\subsection{More Implementation Details}\label{app:hyperparameters}

\begin{figure}
    \vspace{-3ex}
    \centering
    \includegraphics[width=\linewidth]{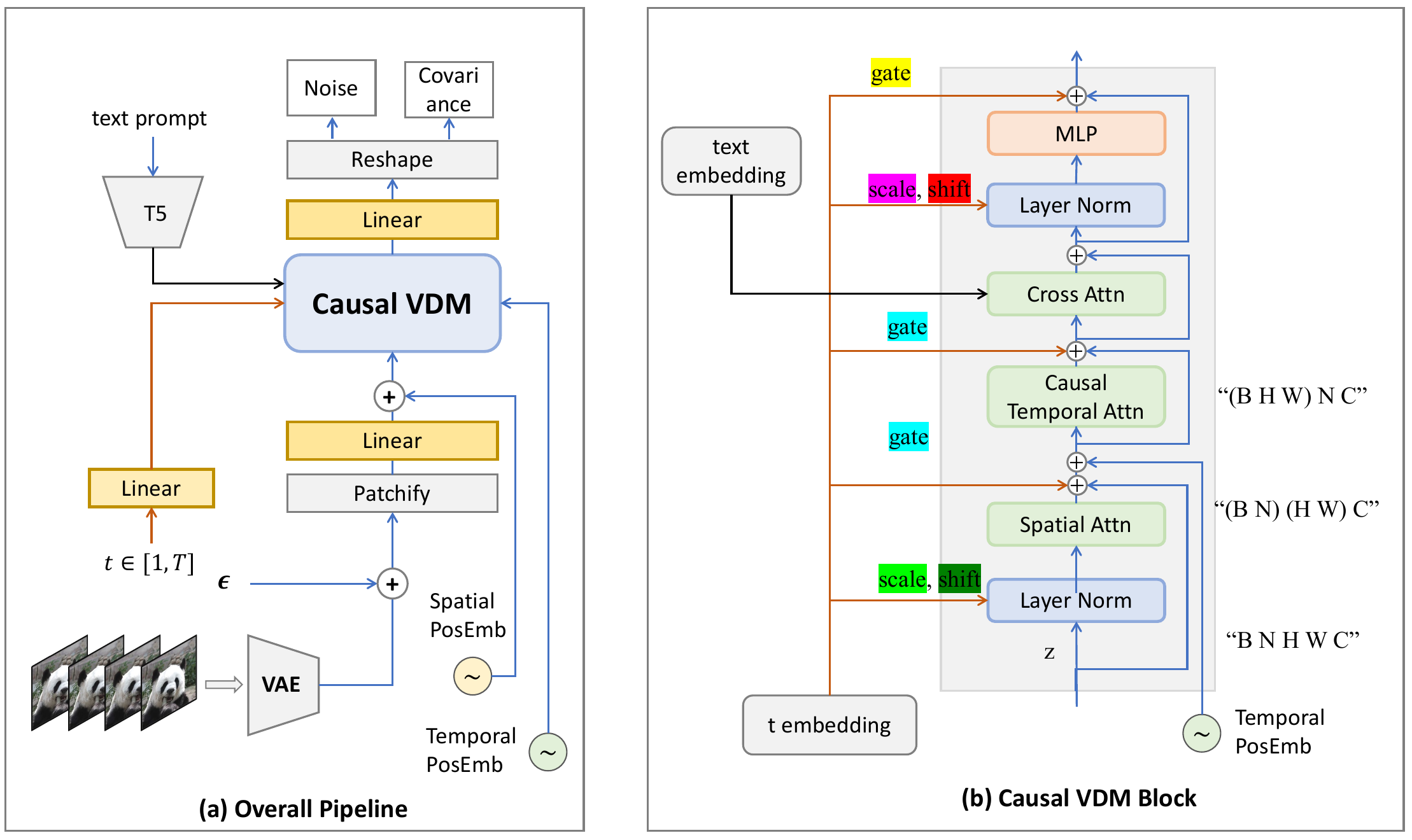}
    \caption{The over all computation pipeline of ViD-GPT (a), and the Causal 
 VDM block (b). The timestep embeddings are input to each block, and devided 6 values, which are served as the scale and shift for adaptive layer normalization, and the gate vaules for MLP layers and attention layers (one color for each).}\label{fig:appmodel} %
    \vspace{-2ex}
\end{figure}

\subsubsection{Model Structure}
We illustrate the overall model structure used in ViD-GPT as shown in Figure~\ref{fig:appmodel}~(a). The causal VDM is stacked by 28 identical blocks as each on shown in Figure~\ref{fig:appmodel}~(b). Following PixArt-$\alpha$~\cite{chen2024pixart}, we use T5 large (\ie, Flan-T5-XXL)~\cite{raffel2020exploring} as the text encoder and use the pretrained VAE from StableDiffusion~\cite{rombach2022high} as the image encoder. 
\subsubsection{Training Details}

Our model is trained over all Transformer parameters with frozen text encoder and VAE encoder. The training consists two stages. We first train the causal modeling ability of ViD-GPT without using frame prompt, on videos with resolution $32\times 256 \times 256$. Then we use longer videos of $65\times 256 \times 256$ to train the model with frame prompt, enabling its autoregressive generation ability. we set the autoregression chunk length (\ie, $n$) as 16 frames, and random keep unnoised frames as prompt with lengths according to the multiples of $n$, \ie, sampling from [1, 17, 33, 49]. The training data is sampled with a frame interval of 3, and shorter videos are filtered out, \eg, in the second stage, videos shorter than $3\times 65=195$ frames are filtered out. During training, we randomly drop out the text prompt with a probability of 0.1 to enable classifier-free guidance~\cite{ho2022classifier}. We use the default DDPM~\cite{ho2020denoising} schedule with $T=1000$, $\beta_1=10^{-4}$, and $\beta_T=0.02$. Our model is trained using AdamW~\cite{loshchilov2018decoupled} optimizer with a learning rate of 2e-5. 
In the first stage, the model is trained with a batch size of 288  for 32k steps. In the second stage, it is trained with a batch size of 144 for 21k steps. 
 
At the inference stage, the length of each autoregression chunk is set as $n=16$. The maximum length of kv-cache is set as $L=49$. We use DDIM sampling schedule~\cite{song2021denoising} with 100 steps, and set the class-free guidance scale as 7.5.

%

\subsection{Details of FVD Evaluations}\label{app:eval}

Fr\'echet Video Distance (FVD)~\cite{unterthiner2019fvd} measures the similarity between generated and real videos based on the distributions on the feature space. We follow prior works~\cite{blattmann2023align,ge2022long,ren2024consisti2v} to use a pretained I3D model\footnote{\url{https://github.com/songweige/TATS/blob/main/tats/fvd/i3d_pretrained_400.pt}} to extract the features, and use the codebase\footnote{\url{https://github.com/universome/stylegan-v}} from StyleGAN-V~\cite{skorokhodov2021stylegan} to compute FVD statistics.

\subsection{Consecutive Frames of Qualitative Examples}\label{app:qualitive}

\begin{figure}
    \vspace{-3ex}
    \centering
    \includegraphics[width=\linewidth]{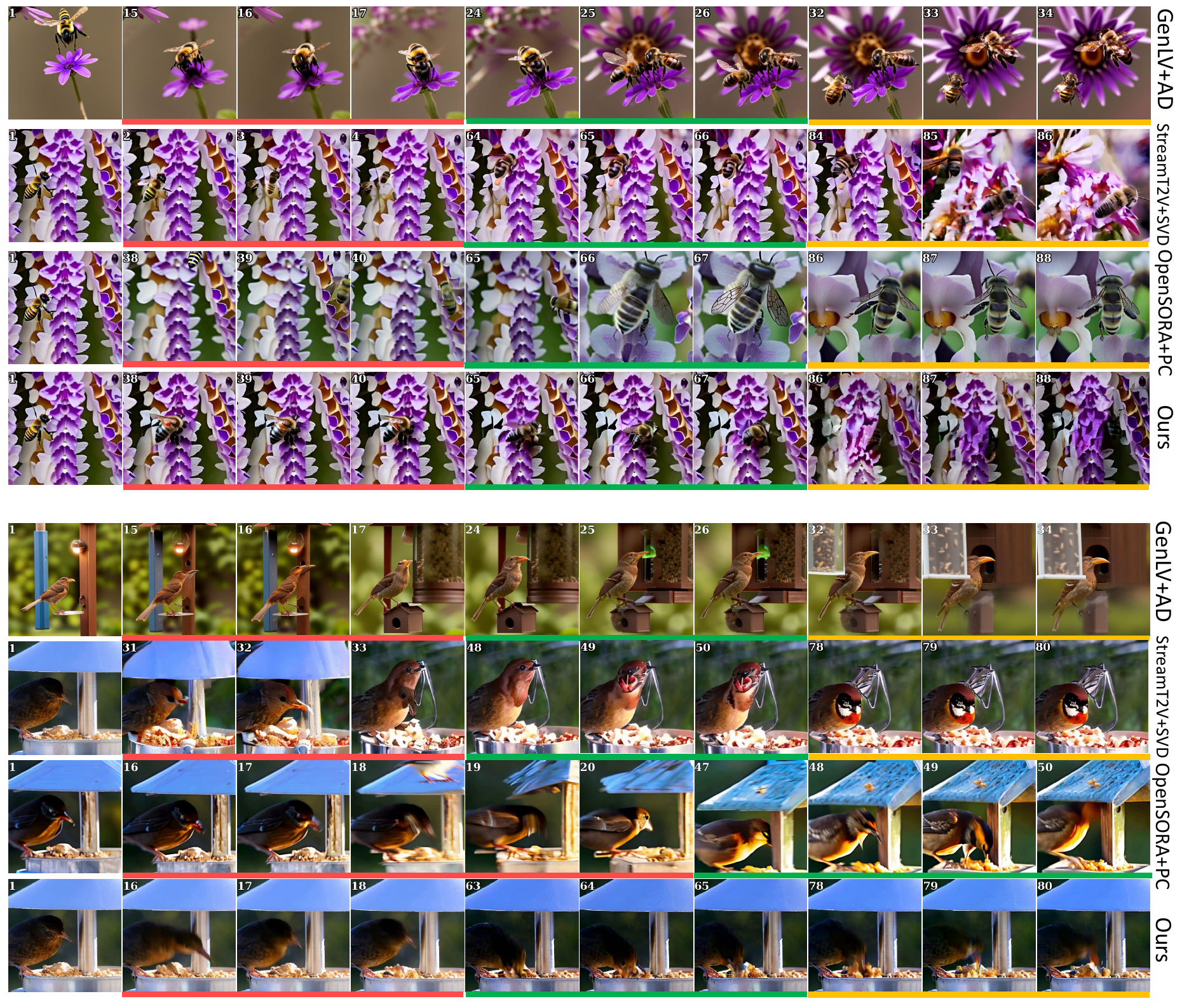}
    \caption{Consecutive frames of qualitative examples from GenLV~\cite{wang2023gen}, StreamT2V~\cite{henschel2024streamingt2v}, OpenSORA~\cite{opensora}, and our ViD-GPT. We select some representative consecutive clips and indicate them with underlines of the same color.}\label{fig:qualitativeApp} %
    \vspace{-2ex}
\end{figure}

Based on the results in Figure~\ref{fig:qualitativeApp}, We have following obersivations: 1) Our ViD-GPT shows better transition consistency comapred to GenLV and StreamT2V. Our generation results have no severe content mutations compared to theirs (\eg, 24$\sim$25-th frames of GenLV and 65$\sim$66-th frames of StreamT2V in Figure~2~(a)). 2) Our ViD-GPT has better long-term consistency compared to these baselines. For example, the purple flower in the result of OpenSORA-FP strats changing at the 85-th frame, and the bird feeder changes after 20-th frame in the result of StreamT2V.






\end{document}